\documentclass[acmsmall]{acmart}
\usepackage{tikz}
\usepackage{graphicx}
\usepackage[edges]{forest}
\usepackage{subcaption}

\usepackage[table]{xcolor}

\newcommand{\commentisunseen}{0}

\usepackage{graphicx}%
\usepackage{multirow}%
\usepackage{amsmath,amsfonts}%
\usepackage{amsthm}%
\usepackage{xcolor}%
\usepackage{textcomp}%
\usepackage{booktabs}%
\usepackage{makecell}%
\usepackage{array}%
\usepackage{adjustbox}%
\usepackage{colortbl}

\definecolor{myred}{RGB}{246, 204, 204}
\definecolor{myblue}{RGB}{164, 194, 244}
\definecolor{mygray}{RGB}{185, 185, 185}
\definecolor{myyellow}{RGB}{255, 221, 179}
\definecolor{mypurple}{RGB}{213, 194, 234}   
\definecolor{mycyan}{RGB}{179, 226, 226}     
\definecolor{myorange}{RGB}{255, 196, 153}  
\definecolor{mydeepgray}{RGB}{150, 150, 150} 
\definecolor{mypink}{RGB}{250, 204, 213}      
\definecolor{myolive}{RGB}{212, 220, 182}    
\definecolor{myblueline}{RGB}{87, 127, 185}
\definecolor{bluelight1}{RGB}{185, 211, 237}
\definecolor{bluelight2}{RGB}{213, 222, 239}
\definecolor{mygreen}{RGB}{168, 209, 201}
\definecolor{greenlight}{RGB}{220, 235, 234}
\definecolor{hidden-draw}{RGB}{177, 177, 177}

\definecolor{myultralightgray}{RGB}{240, 240, 240}  
\definecolor{myultralightpink}{RGB}{252, 228, 232}  
\definecolor{myultralightblue}{RGB}{220, 230, 248}  
\definecolor{myultralightgreen}{RGB}{225, 240, 225}  
\definecolor{myultralightyellow}{RGB}{255, 240, 210} 
\definecolor{myultralightpurple}{RGB}{235, 228, 242} 
\definecolor{myultralightbrown}{RGB}{240, 225, 210}

\newcommand{\commentp}[1]{
}

\newcommand{\fc}[1]{
\ifthenelse{\equal{\commentisunseen}{0}}{
{\color{blue}#1}}
{#1}
}
\newcommand{\FC}[1]{
\ifthenelse{\equal{\commentisunseen}{0}}{
{\color{blue}FC: #1}}
{}
}

\AtBeginDocument{
  }

\setcopyright{acmlicensed}
\copyrightyear{0000}
\acmYear{0000}
\acmDOI{0000000.0000000}

\acmJournal{JACM}
\acmVolume{00}
\acmNumber{0}
\acmArticle{000}
\acmMonth{0}

\setcopyright{cc}
\setcctype{by}
\acmJournal{CSUR}
\acmYear{2026} \acmVolume{0} \acmNumber{0} \acmArticle{}
\acmMonth{1} \acmPrice{} \acmDOI{00.0000/0000000}

\begin{document}

\title{LLM Evolution as an Industry-Scale Ecosystem: A Lifecycle Perspective on Continual Learning}

\author{Hao Jiang}
\authornote{These authors contributed equally to this work.}
\email{jianghao66@huawei.com}
\affiliation{
  \institution{Huawei Technologies Co., Ltd.}
  \country{China}
}

\author{Enneng Yang}
\authornotemark[1]
\email{ennengyang@gmail.com}
\affiliation{
  \institution{Shenzhen Campus of Sun Yat-sen University}
  \country{China}
}

\author{Guojie Zhu}
\authornotemark[1]
\email{zhuguojie2@huawei.com}
\affiliation{
  \institution{Huawei Technologies Co., Ltd.}
  \country{China}
}

\author{Yibin Chen}
\email{chenyibin4@huawei.com}
\affiliation{
  \institution{Huawei Technologies Co., Ltd.}
  \country{China}
}

\author{Yunkun Xu}
\email{xuyunkun1@huawei.com}
\affiliation{
  \institution{Huawei Technologies Co., Ltd.}
  \country{China}
}

\author{Zifu Kou}
\email{kouzifu@huawei.com}
\affiliation{
  \institution{Huawei Technologies Co., Ltd.}
  \country{China}
}

\author{Jiayi Li}
\email{lijiayi42@huawei.com}
\affiliation{
  \institution{Huawei Technologies Co., Ltd.}
  \country{China}
}

\author{Chong Chen}
\authornote{Corresponding author.}
\email{chenchong55@huawei.com}
\affiliation{
  \institution{Huawei Technologies Co., Ltd.}
  \country{China}
}

\author{Zhao Cao}
\authornotemark[2]
\email{caozhao@ruc.edu.cn}
\affiliation{
  \institution{Renmin University of China}
  \country{China}
}

\author{Li Shen}
\authornotemark[2]
\email{shenli6@mail.sysu.edu.cn}
\affiliation{
  \institution{Shenzhen Campus of Sun Yat-sen University}
  \country{China}
}



\begin{abstract}
Continual learning capability is critical for Industrial LLMs, as deployed models must be continuously updated to meet evolving requirements and environments, rather than repeatedly retrained from scratch. However, most existing research focuses on improvements on static benchmarks, failing to capture real industrial needs. In this survey, we reformulate Industrial Continual Learning (ICL) for LLMs as a closed-loop update-and-release problem in a versioned ecosystem, where updates propagate hierarchically to industrial, application-specific models and LLM-powered applications, with capability inheritance and transfer across versions and model families. From this ecosystem perspective, we identify three core challenges: repeated adaptation erodes model plasticity, foundation-model upgrades break capability inheritance, and long-term sustainability is constrained by deployment requirements. 
We then organize the technical landscape of ICL around five lifecycle design principles: preserving plasticity headroom, treating upgrades as capability transfer, enabling trustworthy continual reinforcement learning, making training recipes self-optimizing, and building accountability as a base layer for long-term iteration.
For each principle, we synthesize representative technical directions.
Finally, we evaluate the maturity of each principle and its technical components via an evidence-based lens, identify key gaps hindering real-world deployment, and outline a practical ICL deployment blueprint and a pathway for feeding industrial realities back into academic research.
\end{abstract}

\maketitle




\section{Introduction}
\label{sec:introduction}

\begin{figure}[t]
\centering
\includegraphics[width=1.0\textwidth]{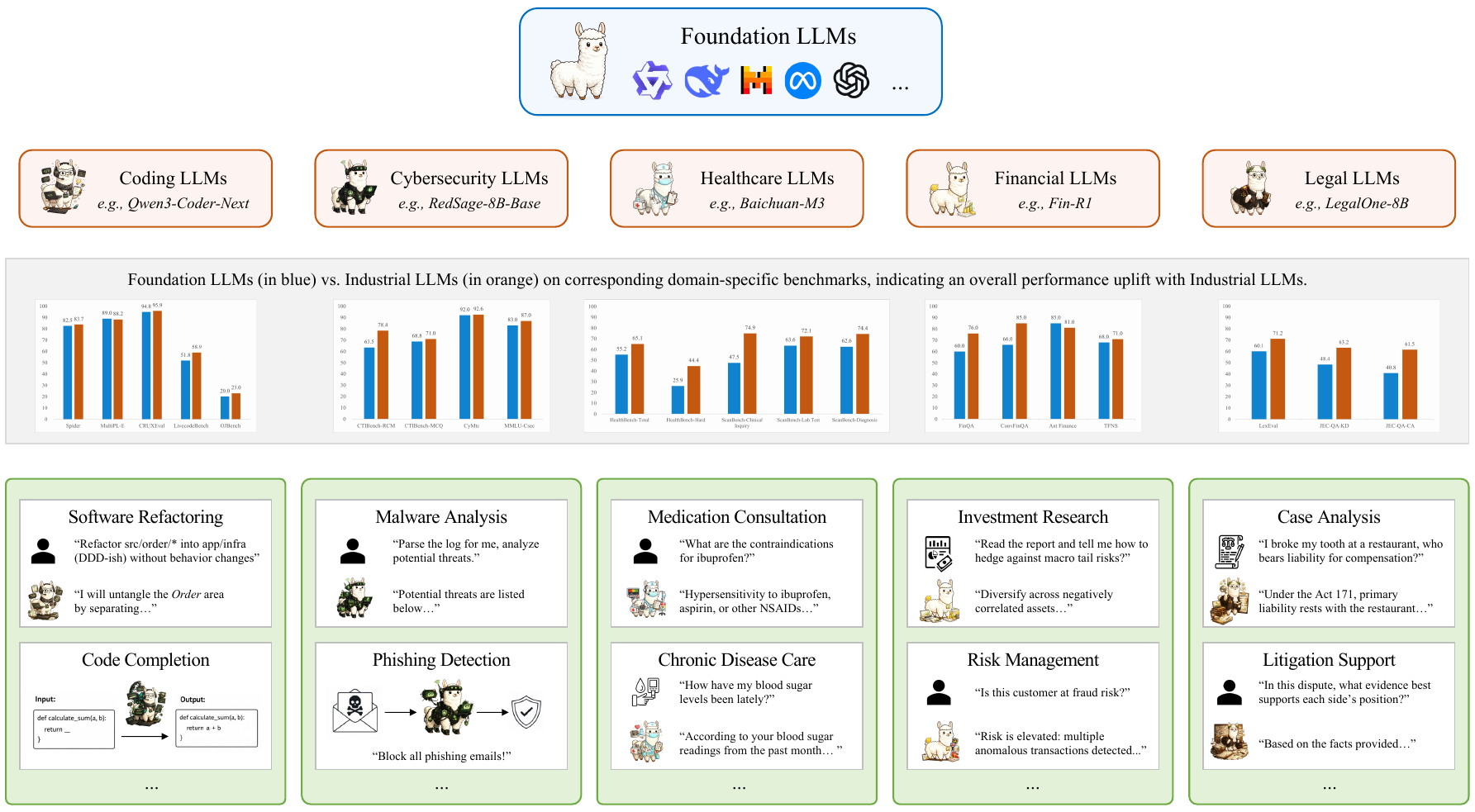}
\caption{A top-down ecosystem of industrial LLMs: from foundation models to domain-specific models, and further to performance comparisons and case studies on industry tasks. The \textit{first row} shows foundation LLMs, serving as the upstream general-purpose backbone; the \textit{second row} shows industrial LLMs built from foundation LLMs through processes such as continual domain tuning, spanning representative verticals (e.g., Qwen3-Coder-Next~\cite{qwen_qwen3_coder_next_tech_report} for coding, RedSage-8B-Base~\cite{suryanto2026redsage} for cybersecurity, Baichuan-M3~\cite{m3team2026baichuanm3modelingclinicalinquiry} for healthcare, Fin-R1~\cite{liu2025finr1largelanguagemodel} for finance, and LegalOne-8B~\cite{li2026legalone} for law), highlighting their domain-tailored capabilities. The \textit{third row} presents bar charts comparing performance on industry benchmarks within each domain: blue bars denote foundation LLMs and orange bars denote industrial LLMs, showing an overall performance uplift of industrial models on most domain tasks. The \textit{fourth row} provides brief case analyses aligned with real industrial needs, such as code completion and software refactoring in coding, and malware analysis and phishing detection in cybersecurity.
}
\label{fig:background}
\end{figure}

\begin{figure}[t]
\centering
\includegraphics[width=1.0\textwidth]{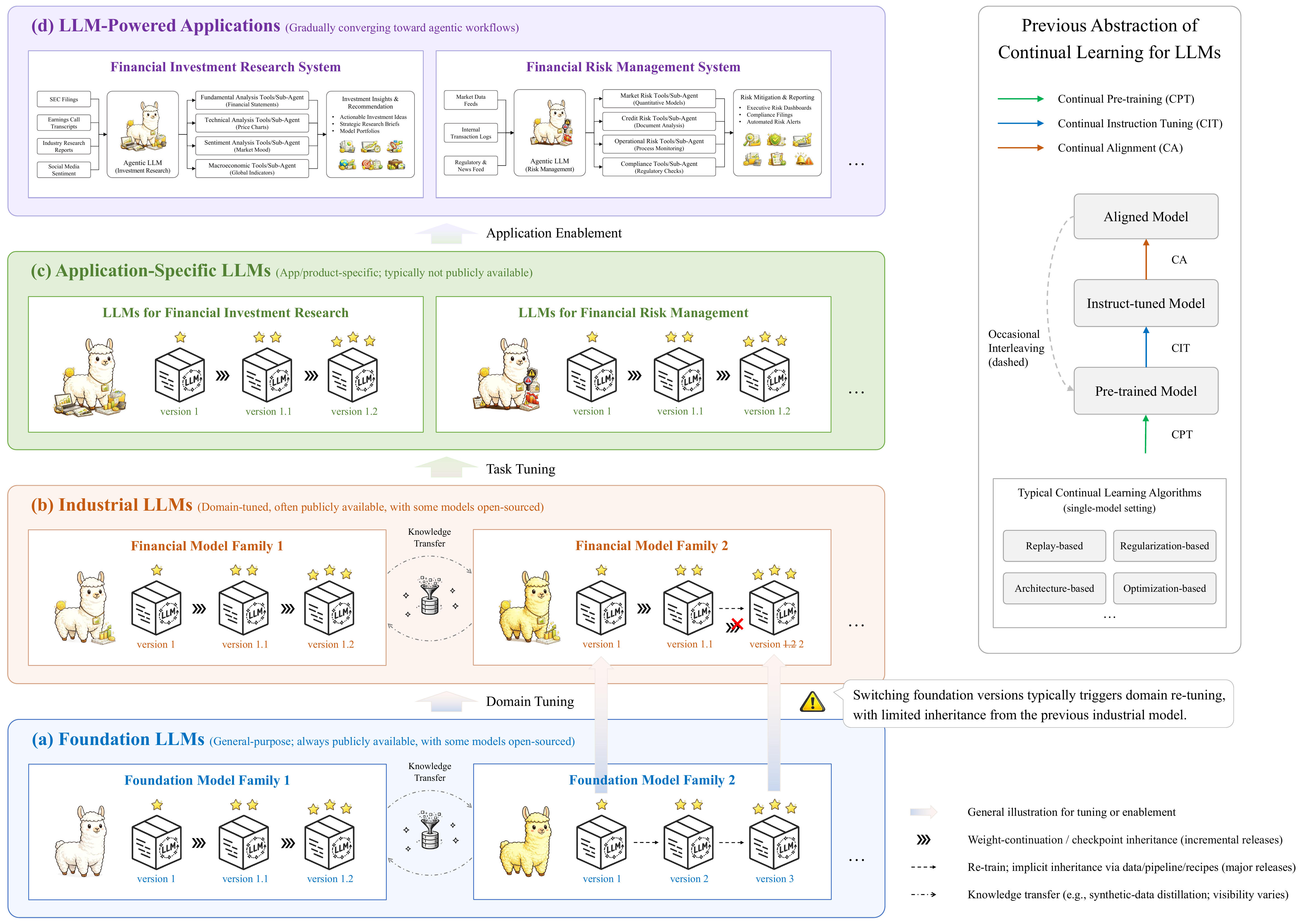}
\caption{\textbf{Left:} Industrial continual learning as a \emph{versioned model ecosystem} rather than a single model cycling through stages. (a) \emph{Foundation LLMs} evolve within multiple families. Within-family upgrades may follow \emph{weight continuation/checkpoint inheritance} (e.g., v1$\rightarrow$v1.1$\rightarrow$v1.2) or \emph{re-training with implicit inheritance} via data, pipelines, and recipes (e.g., v1$\rightarrow$v2$\rightarrow$v3). Cross-family co-evolution can occur via \emph{knowledge transfer}, often through \emph{synthetic-data distillation} (visibility varies). (b) \emph{Industrial LLMs} are domain-tuned descendants of foundation families; switching foundation versions can create inheritance breakpoints and trigger downstream domain re-tuning/refresh cycles. (c) \emph{Application-specific LLMs} further specialize industrial models for app/product targets (task tuning), forming their own version chains. (d) \emph{LLM-powered applications} integrate agentic workflows and tool sub-agents, supported by continual updates across upstream models (application enablement). \textbf{Right:} A common prior abstraction treats continual learning for LLMs as a \emph{single-model}, stage-centric pipeline (CPT$\rightarrow$CIT$\rightarrow$CA)~\cite{wu2024clllm}. While stages may occasionally interleave or re-enter (dashed), the view remains centered on one lineage and single-model CL primitives, obscuring multi-tier, multi-family version chains, heterogeneous upgrade semantics, cross-family knowledge transfer, and upgrade cascades in industrial deployments.}
\label{fig:lifecycle}
\end{figure}

Large language models (LLMs) are becoming the backbone of intelligent services across a wide range of industries. Yet real-world deployment of general-purpose foundation LLMs quickly reveals a key fact: general capability does not automatically translate into reliable, compliant, and cost-bounded performance on domain-critical tasks. Industrial scenarios are often highly specialized, and relying solely on the broad competence of foundation LLMs is frequently insufficient to meet expectations on critical tasks. Fig.~\ref{fig:background} illustrates this point from a top-down ecosystem perspective. Across verticals such as coding, cybersecurity, healthcare, finance, and law, industrial LLMs are not merely application-layer wrappers, but domain experts shaped from foundation LLMs through continual domain training and alignment. More importantly, the bar-chart comparison in the third row of Fig.~\ref{fig:background} suggests that industrial models (orange) often outperform foundation models (blue) on benchmarks within the same domain. Such gains typically stem from better matching domain data distributions, reshaping task instructions and objectives, and co-adapting to domain-specific constraints. Therefore, developing industrial LLMs raises a research and engineering question: how to sustain domain-critical performance under compliance, latency, and cost constraints while models are updated repeatedly over time.

Once LLMs are deployed in real industrial settings, they enter a continuous evolution trajectory: they must repeatedly absorb new information and adapt to emerging task forms over multiple iterations. In practice, this evolution often follows the progressive path depicted on the left of Fig.~\ref{fig:lifecycle}: starting with an industrial model obtained via domain tuning of a foundation model, then further fine-tuning on concrete industry task data to produce application-specific LLMs, and finally integrating with tools, workflows, and interaction systems to form LLM-powered applications. As a result, the question naturally shifts from how to train an industrial model to how to maintain and continually improve it over the long run—which is exactly what this paper emphasizes as Industrial Continual Learning (ICL). Fig.~\ref{fig:lifecycle} characterizes this distinction from a lifecycle perspective: traditional continual learning~\cite{van2022three,wang2024comprehensive} is commonly abstracted as training a single model over a sequence of tasks, focusing on reducing forgetting when learning new tasks, with evaluation typically centered on offline benchmarks and one-shot training processes (right side of Fig.~\ref{fig:lifecycle}). In contrast, the ICL framework highlighted on the left side of Fig.~\ref{fig:lifecycle} focuses on continuous, cross-iteration and cross-version updates.

From an ecosystem perspective, this survey highlights three long-term challenges facing ICL. First, \textit{repeated adaptation reduces plasticity}: continual multi-stage updates accumulate interference and gradually make representations rigid, so the model becomes less able to absorb new knowledge and cannot evolve well over the long run ({C1}). Second, \textit{upgrades break capability inheritance}: practitioners either stay on an old foundation model and keep updating to preserve domain capabilities, but are limited by the old version's ceiling, or they move to a new foundation model to gain stronger general capability, but must rebuild domain and task capabilities again, leading to repeated resource spending ({C2}). 
Third, \textit{updates are constrained by privacy, compute, cost, and latency,} which limits what we can observe and often misaligns training signals with deployment goals  ({C3}).

To address these challenges, this survey redefines ICL as sustainable lifecycle governance, rather than merely a more powerful standalone algorithm.
To support this paradigm shift, unlike an exhaustive catalog of continual learning algorithms, we organize this survey around five lifecycle design principles.
{Principle 1} optimizes not only for current performance but also for measurable plasticity headroom; {Principle 2} treats upgrades as capability transfer rather than checkpoint continuity; {Principle 3} frames RL as long-horizon learning under noisy feedback with explicit safeguards against drift; {Principle 4} makes training recipes adaptive to distribution shift and budget constraints; and {Principle 5} standardizes release criteria and evidence so updates remain comparable, reviewable, and reversible.
Building on these principles, this survey proposes a closed-loop workflow: monitoring motivates each update, transfer and training choices are made explicitly, evaluation is comparable across versions, and release is gated by regression risk with rollback readiness. Within this loop, portability comes from reusable capability representations—such as deltas, adapters, distilled data, and versioned evaluations—so that upgrades become systematic capability transfer rather than repeated retraining. In addition, an accountability mechanism complements this foundation by maintaining auditable evidence logs and decision interfaces that record what changed, when and why it changed, who approved the change, and how it affected downstream behaviors.

Finally, we evaluate over 20 existing industrial LLMs across five domains using an evidence-based approach. We find that the maturity of the five core design principles (and their technical components) varies widely in current industrial LLM practice, and they are rarely packaged as reusable update workflows. 
Based on this maturity review, we further identify structural gaps between research and deployment: which components are close to being directly reusable, which remain effective in academic settings but lack sufficient system-level evidence, and which capabilities are essential in industrial systems yet have been consistently undervalued in the research literature. Building on these gaps, we propose a practical technical blueprint: we break down each ICL principle into concrete development paths, clarifying the relevant scenarios, mechanisms, and verifiable artifacts for each principle, thus providing industrial teams with an actionable roadmap toward a long-horizon, sustainable iteration system. In addition, grounded in the current state of industry practice, we outline a set of practical and high-value directions to inform future academic research.

In conclusion, we highlight our main \textbf{contributions} as follows:
\begin{itemize}
    \item We reformulate ICL as a lifecycle-oriented update-and-release problem in a versioned industrial LLM ecosystem.
    \item We identify three lifecycle challenges in ICL: plasticity erosion, broken capability inheritance, and sustainability under deployment constraints.
    \item We organize the technical landscape around five design principles, rather than isolated algorithmic categories.
    \item We provide an evidence-constrained analysis of industrial practices and outline a deployment blueprint and research roadmap for reliable long-horizon LLM evolution.
\end{itemize}

This paper is structured as follows: Section \ref{sec:introduction} introduces the motivation and scope of this survey. 
Section \ref{sec:traditionalcl} reviews traditional continual learning methods. 
Section \ref{sec:challenges} identifies three fundamental challenges in ICL. Section \ref{sec:framework} proposes a comprehensive taxonomy to address the challenges delineated in Section \ref{sec:challenges}. Section \ref{blueprint} conducts an evidence-based analysis of the technical maturity of over 20 LLMs across five distinct domains, culminating in concrete practical guidelines and promising future research directions. Finally, Section \ref{conclusion} concludes the entire paper.

\section{Traditional Continual Learning}
\label{sec:traditionalcl}

Traditional continual learning studies how a model can acquire new knowledge from a sequence of tasks or data distributions while preserving previously learned capabilities. Its central tension is the stability--plasticity trade-off~\cite{robins1995catastrophic,dohare_loss_2024}: a model should remain plastic enough to learn new domains and tasks, but stable enough to avoid severe degradation on prior knowledge and general capabilities~\cite{alssum2025unforgottensafetypreservingsafety,liogeris2025fullparametercontinualpretraininggemma2,song2025alleviatecatastrophicforgettingllms}. This tension becomes especially important for LLMs, whose incremental training processes, such as continual pretraining $\rightarrow$ continual instruction tuning $\rightarrow$ continual alignment in Figure~\ref{fig:lifecycle}, are increasingly used to keep models up-to-date after deployment.

\begin{figure}[t]
\centering
\includegraphics[scale=0.5]{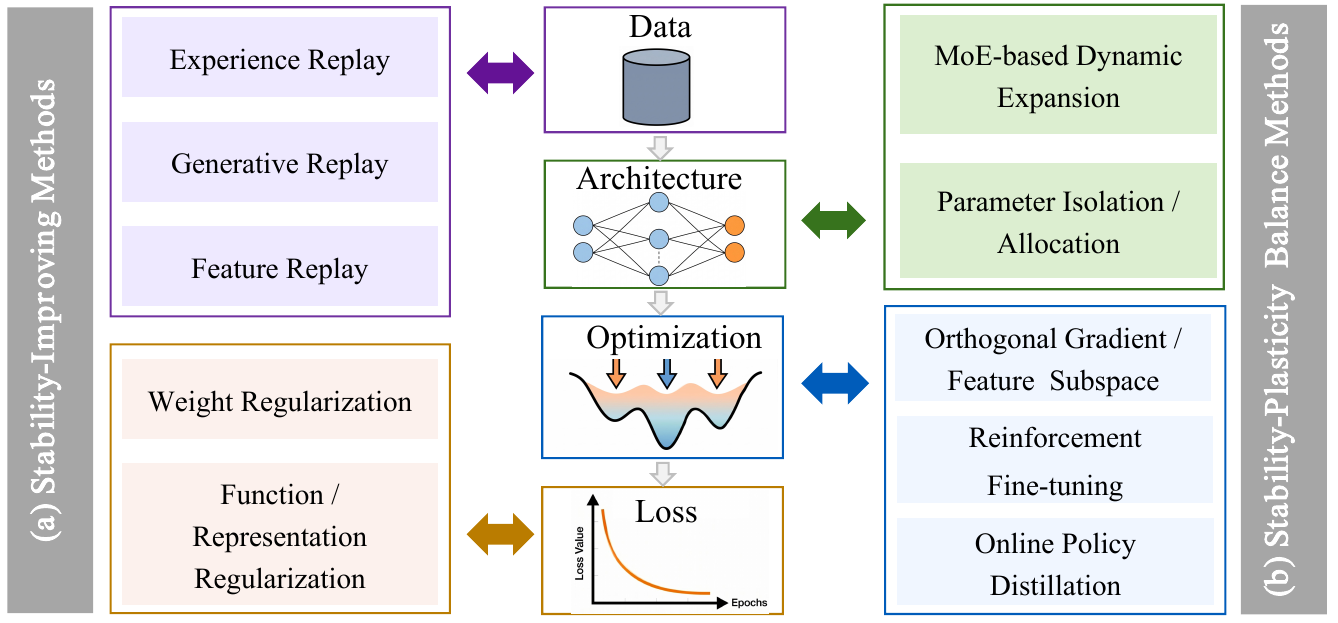}
\vspace{-10pt}
\caption{Several methods for continual learning of LLMs: (a) methods that primarily improve stability and (b) methods that seek a balance between stability and plasticity.}
\label{fig:Continue_Learning_of_LLMs}
\end{figure}

This section organizes existing methods into four categories: replaying a small amount of historical data in new tasks, adding constraints to the objective function, isolating or expanding capacity at the architectural level, and controlling update directions and subspaces through optimization geometry, corresponding to replay-based, regularization-based, architecture-based, and optimization-based methods, respectively~\cite{razdaibiedina2023progressivepromptscontinuallearning,wu2024llamaproprogressivellama,wang-etal-2023-orthogonal}, as shown in
Figure \ref{fig:Continue_Learning_of_LLMs}.

\subsection{Replay-based Methods}
Centered on the stability--plasticity trade-off, replay is crucial in practical incremental-training systems and is one of the most commonly used methods: it mitigates forgetting, maintains coverage of the old distribution, and stabilizes the training process. In other words,
replay anchors learning by interleaving old signals with new updates, but is constrained by storage/privacy and replay fidelity. Various replay strategies differ primarily in the choice of historical data proxies.

Experience replay mixes curated exemplars with new data to improve gradient alignment across increments, at the cost of buffer curation and retention policies~\cite{abbes2025revisiting}. Replay-based methods mainly focus on different data selection strategies~\cite{rebuffi2017icarl,aljundi2019online,bang2021rainbow,chen2024overcoming,wang2024inscl}. For example, iCaRL~\cite{rebuffi2017icarl} selects samples that are closest to the prototype of each class, while MIR~\cite{aljundi2019online} selects samples that are most likely to lie near the decision boundary. Rainbow~\cite{bang2021rainbow} requires the selected memory set to be as diverse as possible.  In the setting of natural language instruction tuning, InsCL~\cite{wang2024inscl} dynamically adjusts the replay ratio according to task similarity, and filters and prioritizes diverse and semantically rich samples based on an instruction informativeness score. HESIT~\cite{chen2024overcoming} selects replay samples by analyzing their gradient influence. The work in~\cite{guo2024efficient} performs multiple rounds of training on high-quality data subsets and dynamically selects the most valuable replay samples according to information gain.

Generative replay synthesizes rehearsal data to avoid storing raw text, trading governance benefits for sample-quality drift and possible hallucinated artifacts~\cite{huang2024mitigating}. For example, LAMOL~\cite{sun2019lamol} introduces generation tokens to guide the generation of pseudo samples during the learning of new tasks, thereby preserving prior knowledge. PCLL~\cite{zhao2022prompt} employs a conditional variational autoencoder to generate task-specific pseudo data from latent representations. SSR~\cite{huang2024mitigating} leverages the in-context learning capability of LLMs to automatically generate synthetic samples and then selects high-quality data through clustering.

The feature-based method replays compressed hidden states or feature statistics to stabilize representations when labeled data is scarce, but requires extra interfaces and cross-version compatibility~\cite{huang2024learning,huang2024etag}. For example, Iscen et al.~\cite{iscen2020memory} introduce a Feature Adaptation Network to adapt previously stored feature vectors to the updated feature space without accessing the corresponding raw training data. Van de Ven et al.~\cite{van2020brain} and BiRT~\cite{jeeveswaran2023birt} propose brain-inspired replay, which replays internal or hidden representations generated by the network’s own context-dependent feedback connections. PASS~\cite{zhu2021prototype} memorizes a representative prototype for each old class and applies prototype augmentation, in the deep feature space to preserve the decision boundaries of previous tasks.

\subsection{Regularization-based Methods}

Regularization-based methods primarily encode the preference for ``preserving prior capabilities'' explicitly into the training objective.
The core idea of weight regularization is to treat the model's existing capabilities as a form of ``prior'', estimate which parameters (or parameter directions) are more critical to those prior capabilities, and then impose stronger penalties on the drift of these critical components when incrementally training on new data. In this way, ``stability'' becomes an explicit term in the optimization objective~\cite{liogeris2025fullparametercontinualpretraininggemma2,song2025alleviatecatastrophicforgettingllms}. In the LLM setting, this line of thinking can be applied both to full-parameter continual pretraining (e.g., in experiments that continue pretraining Gemma2, EWC-style constraints can mitigate degradation of prior capabilities) and to LoRA/low-rank incremental training, where importance-aware constraints are introduced on the low-rank updates to suppress cross-task interference~\cite{liogeris2025fullparametercontinualpretraininggemma2,10.1145/3746252.3761531}. Furthermore, motivated by the observation that different layers contribute differently to general capabilities, Song et al. ~\cite{song2025alleviatecatastrophicforgettingllms} use layer-wise or element-wise importance estimation with adaptive regularization. This applies stronger constraints to layers that should remain more conservative and allows larger updates in layers that should learn more, more directly supporting the goal of improving the new domain while retaining general competence. It is worth noting that the effectiveness of parameter-space regularization depends on how accurately parameter importance is estimated. In other words, in a high-dimensional, non-convex parameter space, choosing a precise estimation method is a key prerequisite for avoiding miscalibrated constraints.

In contrast, feature (function/representation) regularization shifts the constraint from parameter space to function/representation space: even if the weights change, as long as the output distribution or intermediate representations (hidden states) do not drift uncontrollably along key dimensions, the risk of damaging prior capabilities can be reduced~\cite{ding2024boostinglargelanguagemodels}. In settings such as safety alignment or continual fine-tuning, these methods are often implemented as distillation or consistency terms. The old model acts as a teacher, encouraging the new model's outputs or intermediate representations on the new data to stay consistent, which helps reduce drift in alignment behavior or performance on historical tasks~\cite{alssum2025unforgottensafetypreservingsafety}.

\subsection{Architecture-based Methods}

Parameter isolation or allocation views conflicts in incremental training as arising from shared trainable parameters. It reduces interference by assigning relatively independent trainable modules, such as prompts, adapters, and LoRA modules, to different tasks or training stages~\cite{razdaibiedina2023progressivepromptscontinuallearning,araujo2024learningroutedynamicadapter}. Such structural isolation not only substantially alleviates forgetting in LLM incremental training, but also often improves the attainable performance ceiling in the new domain: the new domain is given dedicated trainable capacity, so the learning process does not have to share the same set of parameters with prior-domain capabilities~\cite{razdaibiedina2023progressivepromptscontinuallearning,10.1016/j.knosys.2024.112750,zeng2025neuralnetworksremembermore}. As the number of tasks/domains increases, modular approaches further introduce routing and composition mechanisms: on the one hand, they maintain isolation during module training to reduce training interference; on the other hand, at inference time they leverage learned composition strategies to achieve better generalization and cross-domain transfer~\cite{araujo2024learningroutedynamicadapter,zeng2025neuralnetworksremembermore}. In addition, in continuous task sequences such as information extraction, allocation strategies with multiple LoRA modules or multiple experts are often used to improve performance on the target task while retaining the base extraction capability~\cite{10.1016/j.knosys.2024.112750}.

MoE-based dynamic expansion obtains new capacity by MoE-ifying the model or incrementally expanding experts. On the one hand, a dense FFN can be decomposed into multiple experts, and the router can be trained or updated with sparse top-k activation. This allows new knowledge to be written mainly into the most suitable experts, improving the performance ceiling in the new domain while mitigating forgetting under a controlled compute budget~\cite{zhu-etal-2024-llama,komatsuzaki2023sparseupcyclingtrainingmixtureofexperts,zhang-etal-2022-moefication}. LLaMA-MoE starts from LLaMA, converts it into an MoE model, and continues pretraining while jointly learning gating and expert specialization~\cite{zhu-etal-2024-llama}. Further, Upcycling/UpIT emphasizes rapidly constructing/expanding experts from dense checkpoints or intermediate points during instruction tuning, and improving scalability and stability via parameter merging and router pre-optimization~\cite{komatsuzaki2023sparseupcyclingtrainingmixtureofexperts,hui2024upcyclinginstructiontuningdense}. On the other hand, these methods can also add new experts directly to models that are already MoE architectures: Chen et al.~\cite{chen2023lifelong} proposes a scalable Lifelong-MoE, which gradually adds experts under a continually shifting pretraining stream and freezes existing experts/gating to preserve prior knowledge; MoExtend, in turn, targets modality/task expansion by inserting new experts into a pretrained MoE model, acquiring new capabilities without fine-tuning the original MoE (or the vision encoder) and thus reducing the risk of forgetting~\cite{chen2023lifelong,zhong2024moextendtuningnewexperts}.

\subsection{Optimization-based Methods}

Optimization-based methods resolve stability-plasticity conflicts through training dynamics, using subspace constraints to improve long-term stability and the efficiency of learning in new domains. From an optimization perspective, the core challenge in incremental training is that gradient updates from the new domain may conflict with directions that support prior tasks, leading to degradation of existing capabilities.

Orthogonal gradient subspace methods adopt an optimization-geometric view, interpreting forgetting as the result of conflicts between gradients induced by new-domain data and directions that are critical for previous tasks. Accordingly, they employ gradient projection and orthogonalization to confine parameter updates to subspaces that minimally affect historical directions (or their orthogonal complements), yielding more stable training trajectories over long task sequences. Representative examples include OGD, which performs orthogonal projection directly in parameter space~\cite{farajtabar2019orthogonalgradientdescentcontinual}, and GPM, which extracts and stores bases of historically important gradient subspaces via activation statistics and constrains subsequent updates through projection~\cite{saha2021gradientprojectionmemorycontinual}. To mitigate the loss of plasticity caused by overly restrictive constraints, several works propose adaptive or relaxed projection schemes (e.g., TRGP, SGP, ROGO, and DFGP) that preserve prior knowledge while improving forward transfer and learnability in the new domain~\cite{lin2022trgptrustregiongradient,saha2023continuallearningscaledgradient,yang2023restrictedorthogonalgradientprojection,yang2025revisiting_TPAMI_2025}.

Orthogonal feature subspace methods emphasize explicit disentanglement within trainable representations or low-rank update directions. They assign to different tasks subspaces that are orthogonal (or approximately orthogonal), so that task-specific updates occur along largely independent directions, thereby reducing interference at the representation level. Classical low-rank orthogonal subspace learning constrains each task to a distinct low-dimensional orthogonal subspace and maintains subspace orthogonality via manifold optimization~\cite{chaudhry2020continuallearninglowrankorthogonal}. In parameter-efficient continual learning for LLMs, this principle is often instantiated through LoRA/adapter-based updates; for example, O-LoRA employs orthogonal low-rank adaptation directions to mitigate cross-task interference~\cite{wang-etal-2023-orthogonal}. More recent work further explores approximately orthogonal constructions and adaptive budget allocation (e.g., BiLoRA, OA-Adapter) to improve scalability over long task sequences~\cite{11092988,wan2025adaptivebudgetallocationorthogonalsubspace}.

In addition to the orthogonal projection methods discussed above, several recent studies have explored forgetting mitigation from the perspective of training paradigms. RFT~\cite{lai2025reinforcement} compares two core continual training paradigms, namely supervised fine-tuning (SFT) and reinforcement fine-tuning (RFT), and finds that RFT naturally preserves prior knowledge, whereas SFT leads to catastrophic forgetting of previously learned tasks. Further analysis suggests that an implicit regularization mechanism inherent in RFT is a key contributing factor. Retaining by Doing~\cite{chen2025retaining} shows that the mode-seeking nature of RFT, which stems from its use of on-policy data, helps preserve prior knowledge when learning target tasks, which may reduce forgetting compared with SFT under certain conditions. RL’s Razor~\cite{shenfeld2025rl} analyzes the KL divergence between the fine-tuned policy and the base policy evaluated on the new task, and shows that policy-gradient-based reinforcement learning implicitly favors solutions with minimal KL divergence. Combining the merits of SFT and RFT, CGL~\cite{yao2026cgl} enables rapid adaptation to new tasks and preserves performance stability for old tasks. It further adopts a policy entropy-guided mechanism to dynamically tune the weighting coefficients in their training stages.
Online policy distillation~\citep{song2026survey} provides another related mechanism. During continual adaptation, an old model, a previous policy, or a stronger teacher can provide behavioral targets for the evolving student model. SDFT~\cite{shenfeld2026self} uses a demonstration-conditioned model as its own teacher to generate on-policy training signals, which help preserve prior capabilities while acquiring new skills.

\subsection{From Traditional Continual Learning to ICL}

The above methods provide important algorithmic primitives for incremental LLM adaptation. However, these methods are still largely developed under a single-model abstraction: a model receives new data or tasks over time, and the main objective is to learn the new distribution while preserving previously acquired capabilities.
As shown in Table~\ref{tab:traditional_cl_vs_icl}, ICL for LLMs extends this setting in several fundamental ways. First, industrial LLMs are not isolated checkpoints, but members of a versioned ecosystem in which foundation models, industrial models, application-specific models, and LLM-powered applications co-evolve. Second, model updates are not limited to sequential adaptation on new data; they also include foundation-model upgrades, domain re-tuning, application-specific specialization, alignment refreshes, and release decisions. Third, the success of an update is not measured only by average task accuracy or forgetting, but also by capability inheritance across versions, regression risks on critical slices, deployment cost, latency, compliance, reproducibility, and rollback readiness.
Therefore, traditional continual learning methods are necessary but not sufficient for industrial LLM evolution. These gaps motivate a lifecycle-oriented view of ICL.

\begin{table}[t]
\centering
\small
\caption{Summary of the key differences between traditional continual learning and industrial continual learning (ICL) for LLMs.}
\label{tab:traditional_cl_vs_icl}
\resizebox{\linewidth}{!}{
\begin{tabular}{p{0.18\linewidth} p{0.34\linewidth} p{0.38\linewidth}}
\toprule
\textbf{Aspect} & \textbf{Traditional Continual Learning} & \textbf{Industrial Continual Learning} \\
\midrule
Learning Unit 
& A single model learns a sequence of tasks or data distributions. 
& A versioned ecosystem evolves across foundation models, industrial models, application-specific models, and LLM-powered applications (Fig.~\ref{fig:lifecycle}). \\
\midrule
Update Process 
& Usually follows a linear task sequence with offline training and evaluation. 
& Involves repeated foundation-model upgrades, domain re-tuning, alignment updates, application adaptation, and release decisions (Fig.~\ref{fig:framework}). \\
\midrule
Main Objective 
& Learn new tasks while mitigating forgetting of previous tasks. 
& Sustain long-term capability growth while preserving plasticity, capability inheritance, deployment efficiency and security. \\

\midrule
Evaluation Focus 
& Typically measured by average accuracy, backward transfer, and forward transfer. 
& Requires regression tracking, inheritance tests, safety evaluation, cost measurement, and rollback preparation. \\
\midrule
System Constraints 
& Mainly considers algorithmic performance and training-time resource constraints. 
& Operates under privacy, compliance, noisy feedback, compute budget, latency limits, and governance requirements. \\
\bottomrule
\end{tabular}
}
\end{table}

\section{Challenges for Lifecycle-Oriented ICL}
\label{sec:challenges}

In ICL, deployment rarely follows a single model that moves linearly through CPT$\rightarrow$CIT$\rightarrow$CA (right side of Fig.~\ref{fig:lifecycle})~\cite{wu2024clllm}. Instead, it constitutes a versioned ecosystem as illustrated on the left side of Fig.~\ref{fig:lifecycle}: at the bottom lie continuously iterated Foundation LLMs (involving the coexistence of diverse version iteration chains and cross-model families), the middle layer consists of Industrial LLMs that rely on the foundation models, the upper layer comprises Application-specific LLMs tailored for concrete products, which are ultimately embedded into the top-level agentic LLM application system. 
Because updates propagate vertically across this stack and horizontally across evolving model families, ICL introduces challenges that are not fully captured by classical single-model continual learning. In this section, we organize these challenges into three coupled dimensions.

\subsection{Challenge 1: Plasticity is progressively eroded by multi-stage updates}
\label{challenge1}

Traditional continual learning primarily asks whether a model forgets previously learned knowledge, and proposes a range of remedies such as data replay, parameter regularization, architectural expansion, and gradient-based optimization~\cite{wang2024comprehensive,wang2024comprehensiveV2}. These methods provide useful primitives for stabilizing adaptation, but ICL exposes a broader bottleneck: the more fundamental issue is not forgetting after a single update, but the sustained depletion of model plasticity across a multi-stage update chain. Foundation models are typically trained to maximize ``out-of-the-box'' general capability, with public benchmarks and leaderboard performance serving as the dominant optimization signals~\cite{yang2025qwen3,liu2025deepseek,meta2025llama,zeng2025glm,kimiteam2026kimik25visualagentic}; these objectives rarely account explicitly for a model’s re-adaptability and long-term performance under repeated downstream fine-tuning and along community/industrial derivative lineages. As a result, models in ICL must be repeatedly retrained and realigned along the path (a)–(d) on the left of Fig.~\ref{fig:lifecycle}, yet the base model is often not provisioned with sufficient plasticity budget for such long-term, multi-stage adaptation—thereby turning forgetting into a bottleneck of plasticity exhaustion.

The immediate consequences of plasticity decay include reduced sample efficiency on new tasks, slower learning, and diminishing marginal gains: achieving the same improvement requires larger update magnitudes and longer training cycles, and training becomes more prone to instability and performance fluctuations. More importantly, this issue propagates along the hierarchical structure on the left of Fig.~\ref{fig:lifecycle}: when the foundation model lacks plasticity, domain and task fine-tuning can only make compensatory adjustments within an increasingly constrained parameter space, which in turn exacerbates the system-level difficulty of long-term evolution.
From a survey perspective, this challenge motivates a broader view of methods that preserve or expand future learnability.

\subsection{Challenge 2: Capability inheritance breaks across model families and upgrades}
\label{challenge2}

In ICL, the foundation model is not a single, fixed ``starting point''. Instead, it exists as multiple model families that evolve in parallel and are updated continuously. As a result, inheriting existing capabilities across families and across versions becomes a hard requirement. If we fail to inherit what we already have, the system effectively pays the same cost again for capabilities that were built before. 
First, Fig.~\ref{fig:lifecycle}(a) shows a first type of inheritance demand between Model Family 1 and Model Family 2: \textbf{cross-family transfer}. Different families often emphasize different capability profiles and training goals, so industrial deployment may need to move certain strengths from one family to another. Second, the transition from Fig.~\ref{fig:lifecycle}(a) to (b) shows a second type: \textbf{cross-version transfer}. When Model Family 2 releases a new foundation version, the industry side needs to carry over the domain-specific capabilities and behaviors of existing industrial models onto the new foundation, so that industry capabilities can evolve with foundation upgrades rather than being rebuilt after every upgrade.

Cross-family transfer is hard because model families often differ in architecture, tokenizers, training data, and alignment targets. This makes it difficult to move capabilities via parameter continuity or small, lightweight updates. In practice, transfer often falls back to data-based retraining or distillation-style relearning, which greatly increases the cost of capability transfer. Cross-version inheritance often becomes a hard trade-off. Teams can keep updating an industrial model on the old foundation, but then miss the gains of the new foundation’s capability ceiling. Or they can migrate to the new foundation, but must redo domain and task tuning—so cost and engineering risk accumulate across upgrades. In addition to cost, broken inheritance increases negative flips (i.e., capability regressions) during migration, making upgrades harder to validate and roll out safely.
This challenge calls for methods that make capabilities portable across model versions and families.

\subsection{Challenge 3: Sustainability bottlenecks under real-world deployment constraints}
\label{challenge3}

In the ICL ecosystem, updating LLMs is not a one-time training event. It is a continuous process throughout the system’s lifecycle. At the same time, industrial updates are governed by real-world constraints. First, the update process is often only partly observable: available data is limited by compliance, privacy, access control, and data retention policies, while online feedback can be noisy and delayed. Second, an ICL system has limited control: compute, cost, and latency budgets restrict update frequency, training intensity, and the complexity of deployment strategies. Third, updates must satisfy governance and traceability requirements, such as regression testing, staged rollout, rollback, and reproducibility.

These constraints dominate real LLM deployments. (i) Incomplete, noisy, and delayed feedback means offline supervised signals often fail to reflect real online goals. This creates a persistent gap between ``objectives that are easy to optimize during training'' and ``objectives that must be met in deployment''. As a result, updates increasingly require learning methods that can use interaction feedback directly and optimize long-term utility under constraints. (ii) Any fixed recipe (fixed data mixing, fixed regularization, fixed alignment strength, etc.) will eventually break at some stage during long-term iteration. Relying on expert tuning or combinatorial search is not practical under tight budgets. (iii) When updates propagate across a multi-level stack, regression risks and unclear responsibility become larger: with frequent version changes, non-retrievable data, non-reproducible experiments, or missing evidence, the system cannot clearly answer "who changed what, when and why, or what impact the change had".
Therefore, sustainability requires more than efficient training algorithms. It calls for methods that jointly support feedback-aware learning, automated recipe adaptation, cost- and latency-aware reasoning, knowledge assetization, and accountable release governance.

\section{A Lifecycle-Oriented Taxonomy of ICL Methods}
\label{sec:framework}

In Section~\ref{sec:challenges}, we distilled three challenges that are distinctive to lifecycle-oriented ICL. 
These challenges are often studied in separate literature, but in ICL they interact within the same lifecycle. This section therefore translates the three challenges above into five lifecycle design principles and a closed-loop update-and-release framework (Fig.~\ref{fig:framework}), which together provide the organizing backbone for reviewing the technical landscape.
Our goal is not to introduce another isolated algorithmic taxonomy, but to provide a principle-guided structure for reviewing methods, system practices, evaluation signals, and deployment requirements.

\subsection{Design principles}
\label{principles}

\begin{figure}[t]
\centering
\includegraphics[width=1.0\textwidth]{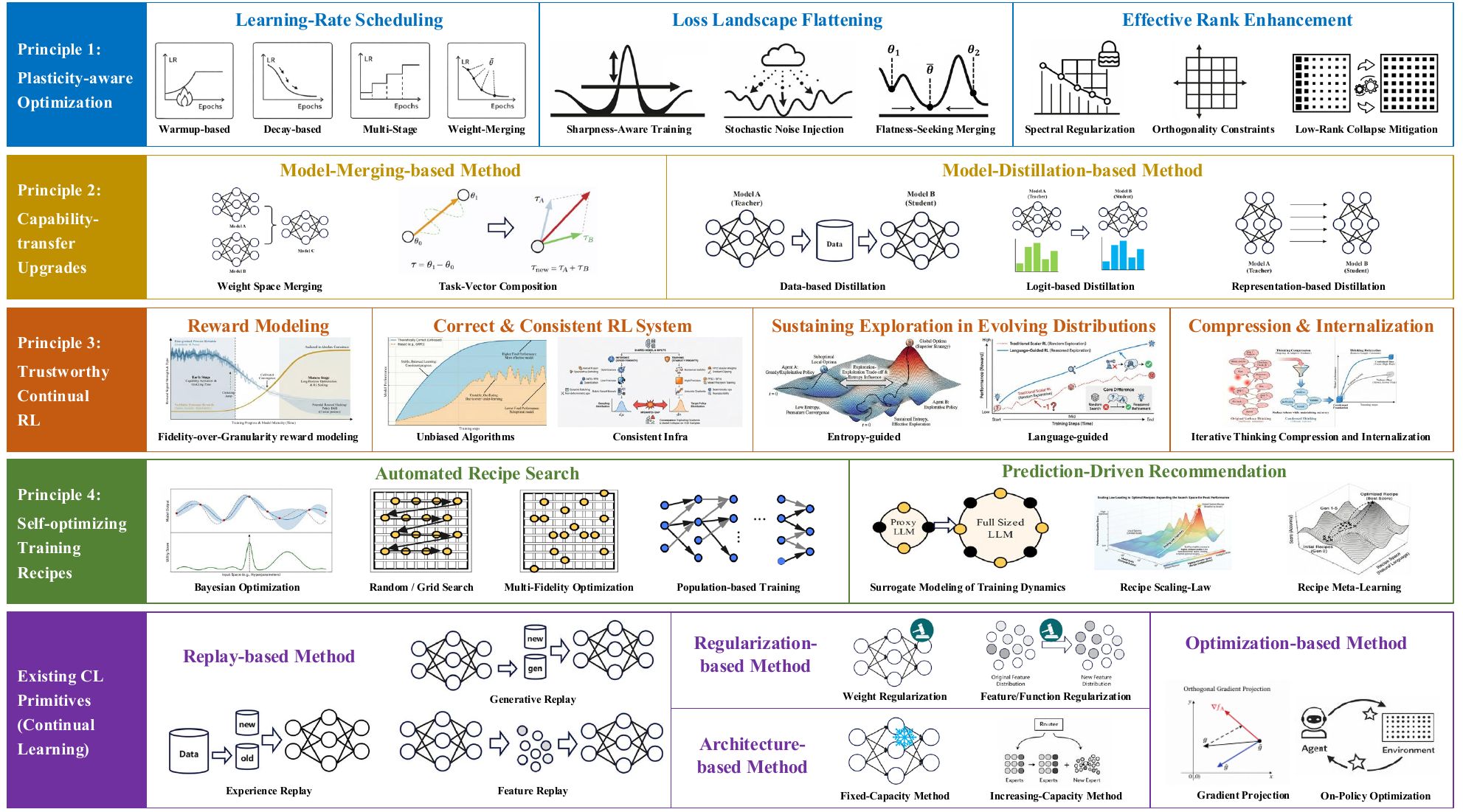}
\caption{\textbf{A taxonomy of supporting technologies} aligned with the four design principles (P1--P4) and existing continual-learning primitives.}
\label{fig:taxonomy}
\end{figure}

In this subsection, we present the five lifecycle design principles in turn. These principles should be understood as system-level contracts rather than isolated techniques: they define what an ICL system must preserve, transfer, optimize, adapt, and document across repeated update cycles. For each principle, we summarize its motivation, representative technical directions (Fig.~\ref{fig:taxonomy}), deployment trade-offs, and evaluation signals. It should be noted that many of the techniques discussed below were not originally developed for ICL. Nevertheless, under the lifecycle view, they become essential building blocks for realizing the corresponding principles in practice.

\subsubsection{Principle 1: Optimize for plasticity headroom, not only current performance}

As discussed in Challenge~1 (Section~\ref{challenge1}), industrial model optimization is often framed as producing a high-performing ``snapshot'', with an implicit assumption that parameters will eventually converge to a stable state and that peak benchmark scores are sufficient to represent model quality~\cite{wang2024comprehensive}. This framing largely holds under static evaluation, but it conflicts with long-term industrial deployment: models in production are updated and specialized repeatedly~\cite{2025Lifelong,dohare_loss_2024}. Therefore, Principle~1 argues that throughout the ICL lifecycle, \textit{future plasticity should be treated as a first-class objective on par with current performance}. Otherwise, over-emphasizing short-term accuracy can shrink the freedom of representations too early, limiting the model’s ability to absorb new data, new tasks, or new domains in later updates~\cite{pmlr-v202-liu23ao}. This principle directly responds to Challenge 1: repeated CPT, CIT, CA, and domain-specific adaptation should not turn model evolution into a one-way process toward reduced learnability.

From a technical perspective, interventions related to future plasticity are already partly visible and reasonably mature, although most of them were not originally designed with lifecycle-level plasticity as the main goal. Concretely, the following three types of techniques can help improve future plasticity:
(i) \textbf{Learning-rate scheduling.} By controlling the magnitude and timing of parameter updates, scheduling can delay premature convergence and slow down the loss of flexibility. Examples include warmup-based strategies~\cite{goyal2017accurate}, decay-based strategies~\cite{loshchilov2018decoupled}, multi-stage schedules~\cite{gururangan2020dontstoppretrainingadapt}, and weight averaging~\cite{izmailov2019averagingweightsleadswider,li2025modelmergingpretraininglarge}. These methods can smooth the optimization path and reduce early rigidity, but plasticity is usually an implicit by-product rather than an explicit target.
(ii) \textbf{Loss-landscape flattening.} These methods shape the optimization landscape to prefer flatter solutions, which empirically helps robustness and transferability. Representative examples include sharpness-aware training~\cite{foret2021sharpnessaware}, noise injection~\cite{neelakantan2015adding}, and averaging methods that favor flat regions~\cite{wortsman2022model,cha2021swad}. However, they often rely on proxy metrics, making it hard to engineer them into stable and consistent lifecycle-level signals.
(iii) \textbf{Effective-rank enhancement.} This line aims to preserve long-term learning potential by maintaining healthy representation capacity, including spectral regularization~\cite{liu2025muonscalablellmtraining}, orthogonality constraints~\cite{farajtabar2019orthogonalgradientdescentcontinual,chaudhry2020continuallearninglowrankorthogonal}, and mitigation of low-rank collapse~\cite{he2025spectralcollapsedrivesloss}. In practice, however, these methods are more like correction or maintenance tools: they modulate representations that have already formed, rather than guiding how representations are formed in the first place. As a result, even though many irreversible representation commitments are made early in training, these methods still have limited control over early-stage decisions.

Overall, while existing techniques have potential value for improving plasticity, from the lifecycle view of Principle~1 they still leave three gaps: (i) future plasticity is not yet an explicit optimization objective; (ii) there is a lack of robust and method-agnostic plasticity signals; and (iii) model selection and early training are still dominated by peak task performance, which prevents lifecycle adaptability from being systematically considered.

\subsubsection{Principle 2: Preserve capability transfer during model upgrades, rather than enforcing weight continuity}

In the standard continual learning formulation, knowledge retention often implicitly relies on \emph{weight continuity}, under the assumption that the next-generation model is simply a fine-tuned descendant of the previous one. However, as discussed in Challenge~2 (Section~\ref{challenge2}), industrial practice is characterized by frequent and heterogeneous foundation-model upgrades, which may involve changes in architecture, tokenizer, or scale. Such changes break continuity and make old checkpoints mathematically incompatible with the new base model~\cite{tan-etal-2025-neural}. Therefore, Principle~2 argues that \textit{model capabilities should be decoupled from specific parameter coordinates and treated as transferable assets}. Rather than hoping domain knowledge remains naturally stored in weights, we should design explicit inheritance mechanisms: extract domain-specific \emph{deltas} or behavioral signatures from the old model and inject them into the upgraded foundation model. The objective shifts from minimizing parameter drift to maximizing high-fidelity transfer of functional capabilities---e.g., domain reasoning skills or compliance constraints---even when the underlying architectures are no longer isomorphic. This principle motivates methods that make capabilities portable across versions and model families.

From a technical perspective, methods originally developed for static model editing, model merging, and knowledge distillation can be repurposed under Principle~2.
(i) For homologous or near-homologous upgrades, \textbf{weight-space model merging} offers an extremely efficient path. Techniques such as task arithmetic~\cite{ilharco2023editing} and TIES-merging~\cite{yadav2023tiesmerging} were originally proposed to combine multiple expert models into a single general model~\cite{Survery_ModelMerging_2024}. In this paper, we argue they can be reframed for capability carryover: domain updates can be packaged as modular \emph{SkillPacks}~\cite{du2025graftllm} and grafted onto a new base model after a version release. However, this direction still has a major gap for heterogeneous upgrades: existing coordinate-alignment methods (e.g., re-basin~\cite{rinaldi2025update}) are often insufficient for large architecture shifts.
(ii) To address this gap, we can rely on \textbf{knowledge distillation}---not as a model compression tool~\cite{Hinton2015DistillingTK}, but as a general mechanism for capability transfer. Concretely, when the weight spaces of the old and new models do not overlap, logit-level alignment~\cite{echterhoff_muscle_2024} and representation-level distillation~\cite{zhan2024over} (originally studied in the student--teacher setting) become key tools to avoid performance degradation
during migration.

Overall, Principle~2 treats cross-version capability inheritance as a recurring requirement that should be supported by repeatable transfer protocols and comparable evaluations. In practice, industrial systems should maintain a capability registry that stores transferable representations (e.g., deltas or distilled data) together with version-comparable tests. With such assets, domain expertise can be preserved even when the underlying foundation model is fully replaced, ensuring that upgrading the general base does not come at the cost of erasing specialized industrial intelligence.

\subsubsection{Principle 3: Optimize RL for trustworthy and sustainable cognitive evolution, not fragile optimization}

As discussed in Challenge~3, ICL operates under many real constraints, such as noisy and delayed online feedback. In this setting, reinforcement learning (RL) becomes a key engine for learning from experience, enabling models to continuously refine their capabilities through open-ended interaction with dynamic business environments and complex workflows. While RL provides a pathway toward ``self-evolving intelligence'', standard RL practice---often centered on maximizing short-term reward---is fundamentally misaligned with ICL. In long-term settings, fragile optimization behaviors (e.g., overfitting to noisy proxy metrics, exploration collapse, or exploiting systematic biases) can not only yield suboptimal performance, but also trigger severe policy degradation and accumulate harmful behavior debt over time. Therefore, Principle~3 calls for shifting RL from fragile reward maximization to stable long-horizon learning: RL should be treated as a control process with explicit defenses against policy drift, reward hacking, and distribution shift, rather than a one-off tuning procedure. The system must carefully balance the trustworthiness of reward signals, the correctness of algorithms and system implementation, and sustained exploration, so that knowledge can accumulate over time without structural collapse.

Technically, the actionable intervention space for this principle spans four interdependent dimensions that collectively enable stable long-horizon RL training:
(i) \textbf{Fidelity-over-granularity reward modeling} acknowledges that reward accuracy and granularity present a trade-off whose optimal balance depends on current model capability, granularity serves as an optimization accelerator, whereas fidelity determines the optimization target~\cite{ye2025correctnessharmonizingprocessoutcome}. In early stages of capability activation, fine-grained process rewards—even those containing heuristic noise—are often indispensable for overcoming sparse signal limitations and accelerating grokking~\cite{wang2024mathshepherdverifyreinforcellms, sun2025rlgrokkingrecipedoes}. However, blind reliance on such dense signals introduces accumulation errors and reward hacking vulnerabilities as training scales~\cite{deepseekai2025deepseekr1incentivizingreasoningcapability, kimiteam2026kimik25visualagentic}. For ICL, the reward architecture should therefore dynamically evolve with model maturity, progressively converging toward verifiable ground truth~\cite{sun2025rlgrokkingrecipedoes, kimiteam2026kimik25visualagentic}. This calibrated convergence, grounded in the conviction that accurate rewards form the critical foundation for RL scaling, ensures that long-term optimization remains anchored to absolute correctness, preventing the systematic policy drift that accumulated reward noise would otherwise introduce.
(ii) \textbf{Correct and consistent RL training systems} are a prerequisite. Any systematic bias that is negligible in one training cycle becomes amplified under continual updates. They require both algorithmic soundness and faithful execution across the training-to-deployment pipeline: at the algorithmic level, training procedures must eliminate systematic biases, including length bias, difficulty bias, expectation estimation bias in advantage calculations, and gradient estimation bias~\cite{liu2025understandingr1zeroliketrainingcritical, zheng2025stabilizingreinforcementlearningllms, yang2026grouprelativeadvantagebiased}. At the infrastructure level, training-inference engine misalignment and MoE router inconsistencies introduce silent policy drift that compounds catastrophically over learning cycles~\cite{ma2025stabilizingmoereinforcementlearning, zheng2025stabilizingreinforcementlearningllms, liu2025speed}. Ensuring both algorithmic correctness and training-inference distribution consistency is therefore critical for ICL, where even minor deviations accumulate into irreversible capability degradation.
(iii) \textbf{Sustaining exploration in evolving distributions} prevents mode collapse and output homogenization that would otherwise terminate the model's capacity to learn from new data, thereby ensuring training stability and longevity across continual updates. Existing methods typically maintain exploration through entropy control~\cite{yu2025dapo, cheng2025reasoningexplorationentropyperspective}, yet high entropy does not necessarily equate to high exploration capability: stochastic diversity without semantic guidance can yield uninformative variability that fails to discover meaningful solution spaces. A promising direction lies in leveraging semantically rich textual feedback for efficient exploration guidance~\cite{hong2026naturallanguageactorcriticscalable, shi2026r3lreflectthenretryreinforcementlearning}.
(iv) \textbf{Iterative thinking compression and internalization} transforms verbose reasoning chains into concise, knowledge-efficient inference paths. This process crystallizes complex deliberation into the model's parameters, maintaining accuracy while significantly increasing reasoning density~\cite{shen2025dast, yu2025dapo}. By internalizing explicit reasoning steps into implicit intuition, the model constructs a more robust cognitive backbone~\cite{team2025kimi_k1_5, team2025kimi_k2, kimiteam2026kimik25visualagentic}. Through iterative compression-relaxation cycles—where strict constraints force knowledge refinement and relaxation phases encourage capability expansion—this consolidated foundation serves as a springboard for exploring novel solution patterns, creating the conditions for sustained capability growth~\cite{wen2025siri, Entropy-Guided}.

While these technical pillars provide the necessary building blocks, they currently function as isolated interventions rather than a cohesive, self-regulating evolutionary system. Principle 3 identifies four structural imperatives for truly sustainable industrial RL: First, reward architectures must progressively converge toward verifiable ground truth to prevent compounding errors and policy drift. Second, algorithmic and systemic correctness requires strict validation, because small biases can compound across cycles and lead to irreversible degradations in behavior. Third, sustained exploration is non-negotiable to prevent mode collapse and maintain adaptive plasticity. Finally, iterative thinking compression-relaxation cycles are critical for capability consolidation and exploratory expansion, driving compounding growth in continual settings. This requires elevating RL from a tactical optimizer to a strategic lifecycle manager that balances the tension between exploiting known paths and exploring the unknown.

\subsubsection{Principle 4: Build self-optimizing training recipes, rather than manual trial-and-error}

Challenge~3 highlights that the ICL paradigm is constrained by compute, cost, and latency.
If every update requires humans to rediscover training configurations via ``trial-and-error'', ICL cannot scale. In current practice, training recipes (e.g., careful choices of data mixing, learning-rate schedules, and regularization strength) are often based on past experience, heuristic rules, or ad-hoc manual tuning~\cite{bengio2012practical, godbole2023deep}. This may be acceptable for one-off training, but it does not scale in the ICL lifecycle: foundation models are upgraded frequently, and data distributions drift, so fixed recipes quickly become outdated and turn into a recurring bottleneck. Therefore, Principle~4 advocates moving from manual tuning to self-optimizing recipes: \textit{the configuration of the training process itself should be produced by an autonomous learning loop, rather than treated as a fixed human artifact}. By viewing the recipe as a dynamic state to be predicted, rather than a set of rules to follow, the system can reduce ``time-to-recipe'' and improve reproducibility under strict compute budgets~\cite{snoek2012practical, hutter2011smac}.

Technically, several lines of automated optimization research---originally developed for more static settings---are promising building blocks for this principle. (i) Recipe optimization has evolved from experience-driven heuristics~\cite{bengio2012practical, godbole2023deep} to \textbf{automated search frameworks} (e.g., Bayesian optimization and Hyperband), which systematize exploration~\cite{snoek2012practical, hutter2011smac}, but are still largely reactive and can be expensive. 
(ii) To support sustainable ICL, we also need \textbf{prediction-driven recommendation} methods, such as neural scaling laws~\cite{kaplan2020scaling, hoffmann2022training} and zero-shot hyperparameter transfer~\cite{yang2022tensor}, although they were originally developed for characterizing static foundation-model pretraining. Concretely, these ideas provide a basis for deriving recipes more systematically: lightweight surrogate models~\cite{akizuki2025surrogate, liu2024regmix} can predict how loss scales with incremental data, enabling lower-cost decisions about data mixtures (e.g., DoReMi~\cite{xie2023doremi}) and training schedules without running full-scale experiments each time. A further trend points to ``LLM-as-Optimizer'', where an LLM uses its prior knowledge to propose and iteratively refine training configurations~\cite{zhang2023using}, forming a closed-loop between model capability and recipe generation.

Overall, a key gap remains: current automation efforts mostly focus on predicting recipes \emph{before} training starts, but lack online adaptivity and cannot adjust recipes based on long-term stability signals~\cite{tribes2023hyperparameter}. Principle~4 therefore calls for \emph{adaptive recipe controllers} as a meta-learning layer: they should be able to detect drift or regression risk on their own and modulate the recipe in real time.

\subsubsection{Principle 5: Build an accountability substrate for long-term updates, not just better updates}

ICL is not a single training run but a sequence of high-stakes updates across tiers, versions, and derived branches, where each change can silently alter capabilities, safety margins, and downstream behaviors. As shown in Challenge~3, in such long-term settings, ``better updates'' are insufficient if the system lacks a stable way to explain, verify, and govern what changed and whether the update should be released. Principle~5 argues that ICL requires a minimum accountability protocol: standardized release criteria, comparable evaluations, and update documentation that make decisions legible and auditable over time.

Concretely, an accountability substrate should provide four primitives.
(i) \textbf{Risk-gated scaling and release criteria.} Update pipelines should explicitly encode capability- or risk-thresholds that trigger escalation, additional safeguards, or no-go decisions, rather than treating deployment as the default end state. Recent frontier governance frameworks operationalize this via thresholded categories and required mitigations, emphasizing measurable capability levels and release gating tied to safeguards and internal decision processes~\cite{openai_preparedness_v2_2025,anthropic_rsp_v2_2_2025,deepmind_fsf_v3_2025}. A risk management framing further motivates that governance is cross-cutting across the lifecycle, and that decision rationales and outcomes should be documented as part of continuous risk control~\cite{nist_ai_rmf_2023}.
(ii) \textbf{Continuous evaluations with an evidence trail.} Because ICL is a series of updates, evaluations must be repeatable, comparable, and traceable across versions (pre- and post-deployment where feasible), with logs and artifacts that support later audit and regression diagnosis. Government-led evaluation programs explicitly position systematic testing as a prerequisite for understanding what new systems can do~\cite{uk_aisi_eval_approach_2024}. Open evaluation infrastructure (e.g., reusable, logged, tool-capable evaluation harnesses) reduces the friction of ``evaluate-before-change'' as the default operational mode~\cite{uk_aisi_inspect_2026}. In addition, frontier safety frameworks increasingly couple capability assessment with explicit plans for updating protocols and documenting disclosures as models evolve~\cite{deepmind_fsf_v3_2025}.
(iii) \textbf{A documentation protocol for model and data deltas.} Accountability requires more than metrics: it requires standardized narrative artifacts that describe intended use, known limitations, evaluation conditions, and what changed in a given update. Model documentation templates (``model cards'') provide a concrete precedent for structured reporting of evaluation context and failure modes~\cite{mitchell_model_cards_2019}. For ICL, the protocol should be \emph{delta-oriented}: each release should include an update note that enumerates data and objective shifts, safety-relevant changes, and evaluated regressions, alongside dataset documentation practices that clarify data motivation, composition, and recommended use~\cite{gebru_datasheets_2021}.
(iv) \textbf{External interfaces for scrutiny and incident learning.} Long-term deployments benefit from mechanisms that enable selective transparency and external input without disclosing sensitive details (e.g., high-level evaluation summaries, red-teaming outcomes, and documented update rationales). Recent scaling and safety frameworks explicitly include transparency components and external engagement as part of their governance loop, while also committing to periodic updates of protocols and disclosures as understanding evolves~\cite{anthropic_rsp_v2_2_2025,deepmind_fsf_v3_2025}.

Principle~5 is intentionally substrate-level: it does not replace technical advances in plasticity, planning, or long-term RL, but makes them \emph{deployable at scale} by turning a complex update stream into a governed, evidence-backed, and communicable process. This substrate also creates a shared interface for the community: comparable update reports, reproducible evaluation traces, and consistent release criteria and decision records that reduce invisible engineering and improve cumulative progress.

\begin{figure}[t]
\centering
\includegraphics[width=1.0\textwidth]{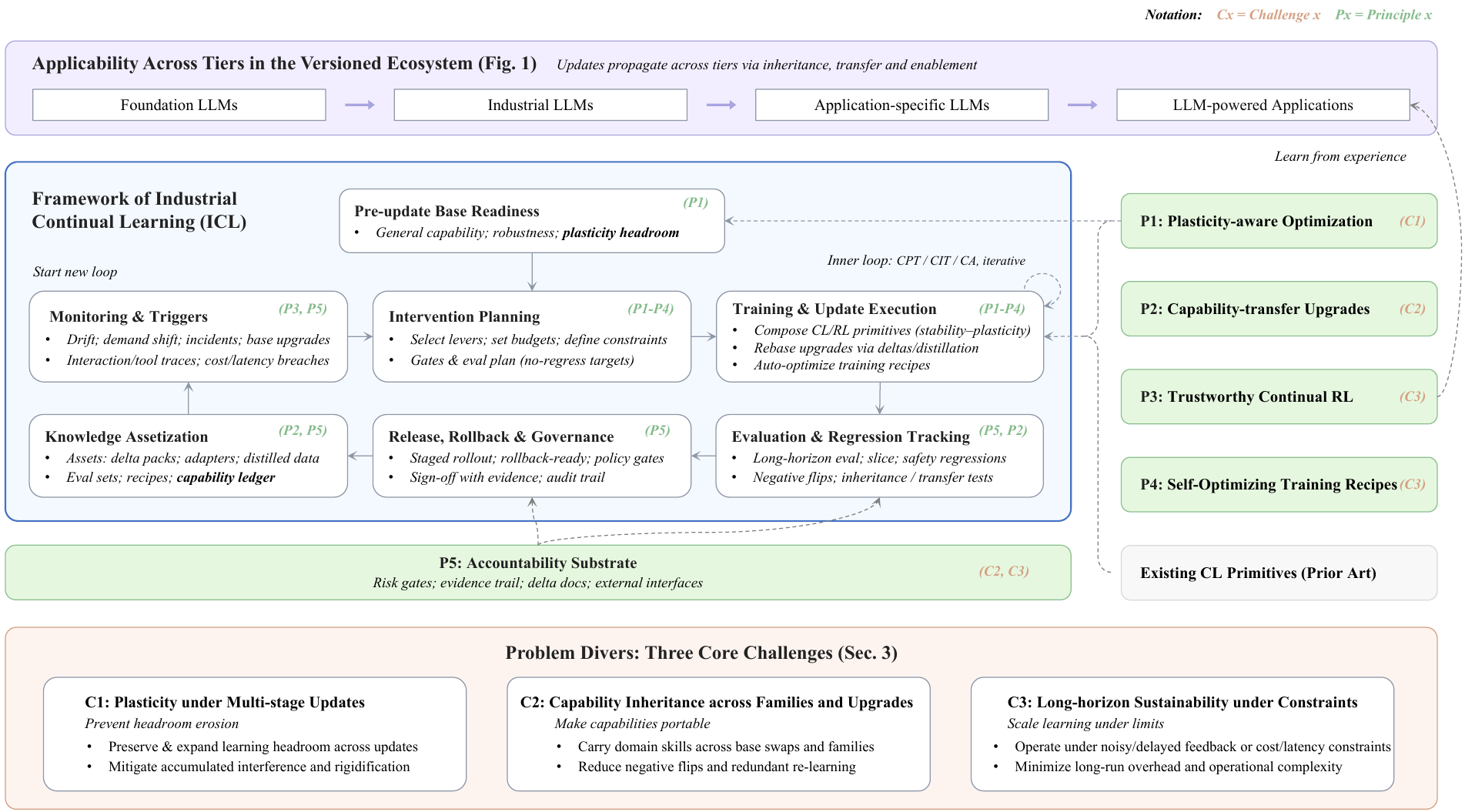}
\caption{\textbf{A closed-loop framework for industrial continual learning (ICL).}
ICL operates as an update-and-release system across tiers in a versioned ecosystem (Foundation $\rightarrow$ Industrial $\rightarrow$ Application-specific models $\rightarrow$ applications; cf. Fig.~\ref{fig:lifecycle}).
Deployment signals trigger iterative cycles of planning, training/update execution (including rebase-and-inherit via deltas/distillation and recipe optimization), evaluation and regression tracking, and release/rollback governance.
Two substrate components---knowledge assetization and accountability---make capabilities portable across upgrades and keep long-term updates auditable.
The framework is organized around three challenges (C1--C3) and five principles (P1--P5) shown on the right and bottom.}
\label{fig:framework}
\end{figure}

\subsection{The closed-loop update-and-release workflow}
\label{sec:workflow}

To connect the five principles above into a system-level view, Fig.~\ref{fig:framework} summarizes ICL as a closed-loop update-and-release workflow.
The framework centers on a repeatable loop with two substrate layers.
An iteration starts from monitoring signals (drift, demand shifts, incidents, base upgrades, and interaction traces) and is translated into an intervention plan under budget constraints and predefined regression criteria.
Training and update execution are the point of action: the system composes reusable CL/RL mechanisms under stability--plasticity considerations, performs rebase-style upgrades via deltas or distillation to preserve domain capabilities across foundation changes, and increasingly treats training recipes as controllable objects to be optimized rather than hand-crafted playbooks.
Each update is then validated through version-comparable evaluations and regression tracking across capability, safety, and cost, followed by staged release with rollback readiness and explicit release criteria.
Two substrate components make the loop scalable across long horizons: knowledge assetization turns deltas, adapters, distilled data, evaluation sets, and recipes into versioned capability artifacts so upgrades do not devolve into ``retrain-and-rediscover'', and accountability provides the evidence trail and decision interfaces that keep iterative updates auditable over time.

This loop is guided by five principles that act as system-level contracts rather than isolated techniques.
P1 treats plasticity headroom as a first-class objective alongside performance, so optimization does not turn into a one-way road toward rigidity (C1).
P2 reframes upgrades around capability transfer, enabling rebase-and-inherit across heterogeneous foundation changes (C2).
P3 elevates learning from experience to a long-term RL discipline that remains trustworthy under continual evolution (C3), while P4 pushes recipe selection toward self-optimizing control to reduce lifecycle risk and operational burden (C3).
Finally, P5 provides an accountability substrate that makes the entire process governable at scale, turning a stream of updates into a legible, evidence-backed workflow.
With this framework in place, the next sections can (i) unpack each principle as a design requirement and (ii) ground the resulting workflow in application-facing expectations and blueprint-level practices.

\section{From Evidence to Blueprint: Observations, Practice Guidelines, and a Research Roadmap}
\label{blueprint}

Section~\ref{sec:framework} introduced the principles required for ICL systems, yet their empirical grounding in real-world practice remains limited. To address this, we first anchor the discussion in publicly observable evidence by presenting a principle-aligned taxonomy of enabling techniques and an evidence-constrained adoption profile based on current reports and artifacts. We then use the framework developed above to sketch brief, illustrative guidelines for designing and evaluating ICL systems, without aiming to exhaust the space of implementation variants. Finally, informed by feedback from industrial practice, we highlight several directions that are both scientifically and practically valuable for the academic community, helping accelerate breakthroughs through iterative advances across academia and industry.

\subsection{Evidence-constrained view of current practice}
\label{subsec:current_status}

\subsubsection{A taxonomy for evidence coding and comparable analysis}
\label{subsubsec:tech_taxonomy}

Section~\ref{principles} has already provided a detailed account of the principles and their technical underpinnings. To turn these principles into comparable and countable units for analysis, Fig.~\ref{fig:taxonomy} presents a structured taxonomy: it is organized by principles (row-wise), summarizes representative technique clusters under each principle, and includes—at the bottom—canonical primitives from single-model continual learning~\cite{abbes2025revisiting,vsliogeris2025elastic,chen2023lifelong}. P5 (the accountability substrate) is comparatively clearer at the protocol level, where practices such as risk-gated releases, evidence-based evaluation trails, and rollback-ready governance have been increasingly recognized as necessary for frontier deployment; therefore, we do not include P5 in Fig.~\ref{fig:taxonomy}. This taxonomy is not intended to exhaustively enumerate all possible mechanisms; rather, it provides a shared conceptual coordinate system and labeling scheme that enables the evidence-driven aggregation in Section~\ref{subsubsec:adoption_profile} based on publicly observable practice. Accordingly, techniques that are not observed in the figure should not be interpreted as absent from industry, but as potentially undisclosed or externally unverifiable given the limits of public reporting.

\subsubsection{What we can observe: an evidence-driven adoption profile}
\label{subsubsec:adoption_profile}

Assessing the current state of practice is challenging because ICL pipelines are typically only \emph{partially observable}. Public disclosures tend to emphasize high-level training objectives, selected components from post-training/alignment, and headline evaluation outcomes, while rarely exposing the full decision logic and control variables that govern iterative updates and releases. Given this constraint, we restrict our analysis to two classes of systems for which comparatively stable public artifacts exist: \emph{industrial models} and \emph{foundation models}. In contrast, \emph{application models} embedded in specific products are difficult to include systematically because disclosures are intermittent and rarely provide sufficient technical resolution; we therefore discuss them only when verifiable public evidence is available.

To avoid over-interpreting ``not disclosed'' as ``not used'', we adopt an evidence-constrained reading: we characterize only those adoption signals that can be directly supported by verifiable public materials, rather than inferring practices that may exist behind proprietary systems. Table~\ref{tab:application-matrix-final} operationalizes this approach by mapping the taxonomy mechanisms to a simple three-level coding scheme. \textbf{Core} denotes mechanisms explicitly described in official technical reports, system cards, or training/alignment process narratives and presented as central to the pipeline; \textbf{Auxiliary} captures mechanisms for which some verifiable but limited or incidental evidence exists, without enough information to establish centrality; and \textbf{Absent} indicates insufficient public evidence---not proof of non-adoption.

Under this evidence standard, public reporting exhibits a pronounced observability bias: mechanisms that sit at the center of external narratives, or that naturally appear in training and alignment descriptions, are more likely to be documented. Conversely, mechanisms that depend on fine-grained implementation details are frequently unobservable in public artifacts due to insufficient disclosure granularity. As a provisional summary, mechanisms aligned with P3 (especially Trustworthy Continual RL) more often produce \textbf{Core}-level observable signals—for example, Qwen3 in the General category and Qwen3-Coder-Next in Code/SE are both marked Core in that column. By contrast, mechanisms more related to P1/P2/P4 are more frequently categorized as \textbf{Absent} in public artifacts (reflected by many ``\textendash'' entries), with Plasticity-aware Optimization being a representative case with scarce recorded evidence. Meanwhile, canonical continual-learning primitives (Existing CL Primitives) tend to appear more often as \textbf{Aux} signals—for instance, CodeGemma is marked Aux in that column. Importantly, these statistics should be interpreted as an evidence-supported adoption profile, not as a lower bound on true industrial adoption rates. Their primary value is to provide an actionable entry point for Section~\ref{subsec:future_expectations}: identifying which mechanisms already have a shared reporting language in public artifacts, and which parts of the pipeline would benefit most from verifiable interfaces and more structured disclosure.

\begin{table*}[t]
\centering
\small
\setlength{\tabcolsep}{3pt}
\renewcommand{\arraystretch}{1.12}
\providecommand{\Core}{}
\providecommand{\Aux}{}
\providecommand{\Abs}{}
\renewcommand{\Core}{\textbf{Core}}
\renewcommand{\Aux}{\textit{Aux}}
\renewcommand{\Abs}{\textcolor{black!35}{\textendash}}
\caption{Adoption of taxonomy-defined methodological interventions across foundation models (general domain) and industrial models, based only on publicly documented evidence.}
\label{tab:application-matrix-final}

\begin{adjustbox}{max width=\textwidth,center}
\centering
\begingroup
\setlength{\aboverulesep}{0pt}
\setlength{\belowrulesep}{0pt}
\begin{tabular}{@{}p{20mm} p{40mm} c c c c >{\columncolor{gray!15}[\tabcolsep][0pt]}c@{}}
\toprule

\textbf{Domain} &
\textbf{Model} &
\makecell{\textbf{Plasticity-aware}\\\textbf{Optimization}} &
\makecell{\textbf{Capability-transfer}\\\textbf{Upgrades}} &
\makecell{\textbf{Trustworthy}\\\textbf{Continual RL}} &
\makecell{\textbf{Self-optimizing}\\\textbf{Training Recipes}} &
\makecell{\textbf{Existing CL Primitives}\\\textbf{(Continual Learning)}} \\
\midrule


\multirow{6}{*}{General} 
& Qwen3~\cite{yang2025qwen3} & \Abs & \Abs & \Core & \Aux & \Abs \\
& DeepSeek-V3.2~\cite{liu2025deepseek} & \Abs & \Abs & \Core & \Abs & \Abs \\
& Llama 4~\cite{meta2025llama} & \Abs & \Abs & \Aux & \Abs & \Abs \\
& Kimi K2.5~\cite{kimiteam2026kimik25visualagentic} & \Abs & \Abs & \Core & \Abs & \Abs \\
& GLM-4.5~\cite{zeng2025glm} & \Abs & \Abs & \Aux & \Abs & \Abs \\
& Grok 4.1~\cite{xai_grok41_modelcard_2025} & \Abs & \Abs & \Aux & \Abs & \Abs \\
\specialrule{0.9pt}{0pt}{0pt}

\multirow{4}{*}{Code/SE}
& Qwen3-Coder-Next~\cite{qwen_qwen3_coder_next_tech_report}
  & \Abs & \Aux & \Core & \Core & \Aux \\
& Seed-Coder~\cite{seed2025seedcoderletcodemodel}
  & \Abs & \Abs & \Core & \Abs & \Abs \\
& DeepSeek-Coder-V2~\cite{deepseekai2024deepseekcoderv2breakingbarrierclosedsource}
  & \Abs & \Abs & \Core & \Abs & \Aux \\
& CodeGemma~\cite{codegemmateam2024codegemmaopencodemodels}
  & \Abs & \Abs & \Aux & \Abs & \Aux \\
\midrule

\multirow{3}{*}{Cybersecurity}
& Sec-LLM-8B-Instruct~\cite{weerawardhena2025llama31foundationaisecurityllm8binstructtechnicalreport}
  & \Abs & \Abs & \Aux & \Abs & \Abs \\
& Llama-Primus-Merged~\cite{yu2025primuspioneeringcollectionopensource}
  & \Abs & \Aux & \Abs & \Aux & \Core \\
& RedSage~\cite{suryanto2026redsage} & \Abs & \Abs & \Abs & \Abs & \Core \\
\midrule

\multirow{5}{*}{Healthcare}
& Baichuan-M3~\cite{m3team2026baichuanm3modelingclinicalinquiry}
  & \Abs & \Aux & \Core & \Abs & \Aux \\
& EHR-R1~\cite{liao2025ehrr1reasoningenhancedfoundationallanguage}
  & \Abs & \Abs & \Core & \Abs & \Abs \\
& QuarkMed~\cite{li2025quarkmedmedicalfoundationmodel}
  & \Abs & \Abs & \Aux & \Aux & \Abs \\
& Me-LLaMA~\cite{xie2024llamafoundationlargelanguage}
  & \Abs & \Abs & \Abs & \Abs & \Core \\
& Med42-v2~\cite{christophe2024med42v2suiteclinicalllms}
  & \Abs & \Abs & \Aux & \Abs & \Abs \\
\midrule

\multirow{5}{*}{Finance}
& FinBloom~\cite{sinha2025finbloomknowledgegroundinglarge}
  & \Abs & \Abs & \Abs & \Abs & \Aux \\
& Fin-R1~\cite{liu2025finr1largelanguagemodel}
  & \Abs & \Abs & \Aux & \Abs & \Abs \\
& DianJin-R1~\cite{zhu2025dianjinr1evaluatingenhancingfinancial}
  & \Abs & \Abs & \Aux & \Abs & \Abs \\
& FinGPT~\cite{yang2025fingptopensourcefinanciallarge}
  & \Abs & \Abs & \Core & \Abs & \Aux \\
& QianfanHuijin~\cite{li2025qianfanhuijintechnicalreportnovel}
  & \Abs & \Abs & \Core & \Abs & \Aux \\
\midrule

\multirow{5}{*}{Legal}
& LegalOne~\cite{li2026legalone} & \Aux & \Abs & \Core & \Abs & \Aux \\
& InternLM-Law~\cite{fei-etal-2025-internlm}
  & \Abs & \Abs & \Abs & \Abs & \Aux \\
& LawGPT~\cite{zhou2024lawgptchineselegalknowledgeenhanced}
  & \Abs & \Abs & \Abs & \Abs & \Aux \\
& SaulLM-54B/141B~\cite{colombo2024saullm54bsaullm141bscaling}
  & \Abs & \Abs & \Aux & \Abs & \Aux \\
& Lawyer~LLaMA~\cite{huang2023lawyerllamatechnicalreport}
  & \Abs & \Abs & \Abs & \Abs & \Aux \\

\bottomrule
\end{tabular}
\endgroup
\end{adjustbox}

\vspace{0.5ex}
\begin{minipage}{\linewidth}
\footnotesize\raggedright
\textbf{Legend:}
\textbf{Core} = explicitly documented as a central pipeline component in public technical artifacts;
\emph{\textbf{Aux}} = verifiable but limited/indirect evidence, with unclear centrality;
\textbf{Abs (\textendash)} = not evidenced in public artifacts (not proof of non-adoption).
\end{minipage}
\end{table*}

\subsection{Principle-aligned guidelines for closed-loop ICL}
\label{subsec:future_expectations}

\begin{figure}[t]
\centering
\includegraphics[width=1.0\textwidth]{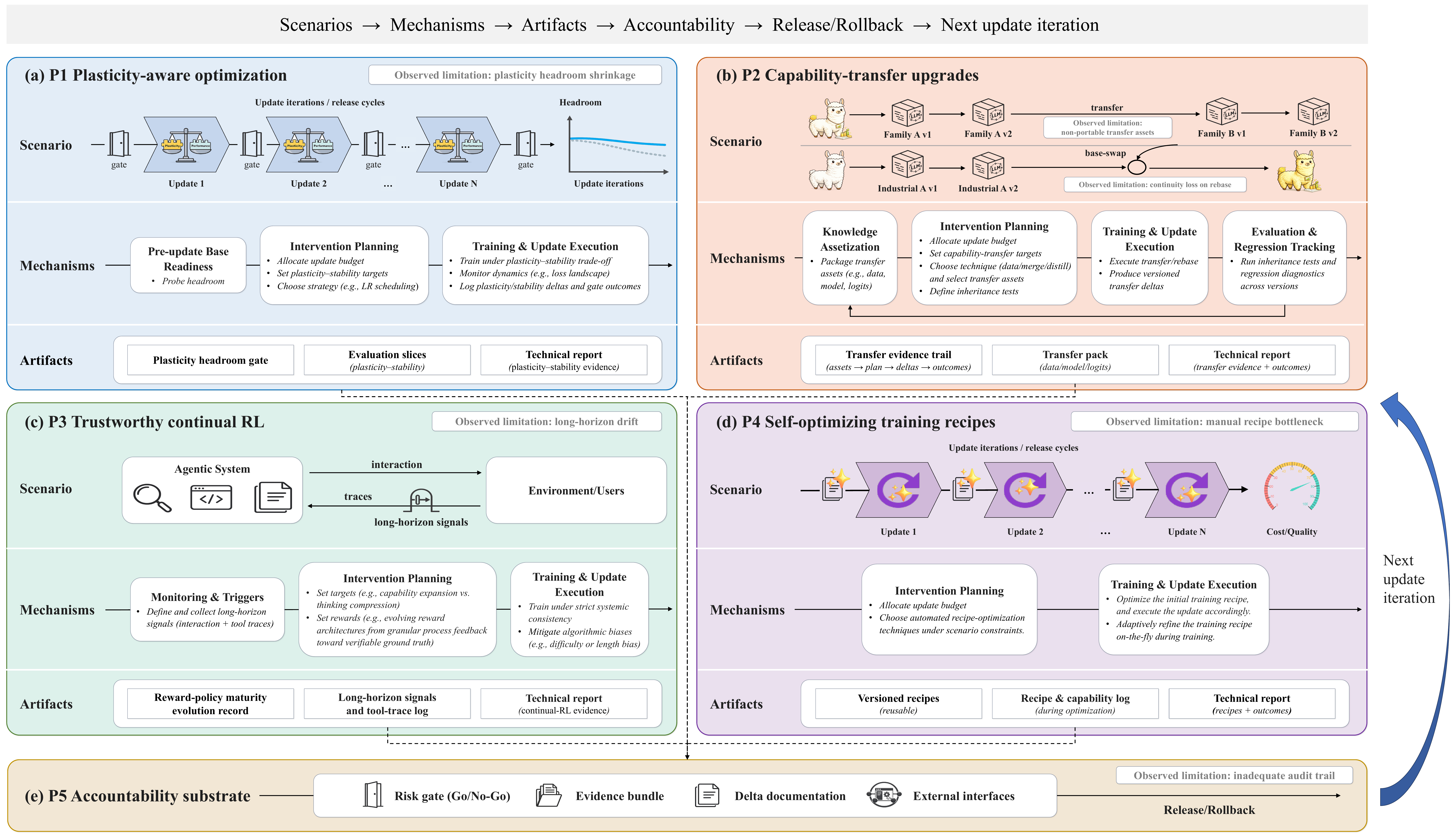}
\caption{Closed-loop update-and-release guidelines for industrial continual learning (ICL), mapping P1--P5 to scenarios, mechanisms, verifiable artifacts, and release/rollback governance across iterations.}
\label{fig:guidelines}
\end{figure}

Building on the framework above, this section links each principle to the scenarios it addresses, the mechanisms that support it, and the artifacts teams can produce and verify (Fig.~\ref{fig:guidelines}). Specifically, for each principle, we describe the scenarios where it arises, the mechanisms that support it in practice, and the key artifacts it produces.
Principles P1–P4 correspond to core system capabilities spanning optimization, capability upgrades, continual alignment for agentic RL, and self-optimizing training recipes, whereas P5 serves as a cross-cutting accountability substrate that links evidence, risk gating, and the release/rollback loop. Together, these principles enable ICL systems to sustain capability inheritance, adapt to evolving distributions, and maintain deployment reliability under continual iteration.

\paragraph{\textbf{Guideline 1}: Headroom-budgeted updates across release cycles}

This guideline is relevant for update iterations across release cycles that are mediated by explicit gates, where the key gap is shrinking headroom over update iterations. Each transition requires operationally tracking how much headroom remains and constraining how much the upcoming update is allowed to consume via an allocated update budget.
As illustrated in Fig.~\ref{fig:guidelines}(a), the mechanism is a three-step control loop: (i) \textit{Pre-update Base Readiness} to run a headroom probe and assess the pre-update state; (ii) \textit{Intervention Planning} to allocate update budget, set plasticity/stability targets, and choose strategy (e.g., LR scheduling); and (iii) \textit{Training \textnormal{\&} Update Execution} to train under the plasticity--stability trade-off, monitor dynamics (e.g., loss landscape), and log plasticity/stability deltas together with gate outcomes. Corresponding artifacts include a \textit{plasticity headroom gate}, \textit{eval slices} (plasticity--stability), and \textit{technical reports including plasticity and stability}. Overall, this guideline makes headroom probing, budget allocation, and gated execution explicit in every update iteration to manage headroom shrinkage under continual updates.

\paragraph{\textbf{Guideline 2}: Capability inheritance across upgrades and rebases}

This guideline targets two common scenarios: (i) cross-family upgrades from Family~A (v1/v2) to Family~B (v1/v2), where the gap is \textit{no portable transfer assets}; and (ii) base-swap/rebase from Industrial~A (v1/v2) to Industrial~B, where the gap is \textit{continuity loss on rebase}. As illustrated in Fig.~\ref{fig:guidelines}(b), the workflow comprises four stages: \textit{Knowledge Assetization} to package transfer assets (e.g., data/model/logits); \textit{Intervention Planning} to allocate update budget, set targets of capability-transfer, choose technique (merge/distill) and select transfer assets, and define inheritance tests; \textit{Training \textnormal{\&} Update Execution} to perform transfer/rebase and produce versioned transfer deltas; and \textit{Evaluation \textnormal{\&} Regression Tracking} to run inheritance tests and regression diagnosis across versions. The corresponding artifacts include a \textit{transfer evidence trail} (from assets to plan, deltas, and outcomes), a \textit{transfer pack} (data/model/logits), and \textit{technical reports, covering the full transfer process}. Overall, the guideline makes capability-transfer upgrades a repeatable, evidence-backed workflow that directly addresses gaps in portability and continuity.

\paragraph{\textbf{Guideline 3}: Trustworthy continual RL under long-horizon feedback}

In the scenario shown in Fig.~\ref{fig:guidelines}(c), an agentic system interacts with the environment/users, producing traces and receiving long-horizon signals. The mechanism follows a three-stage loop: \textit{Monitoring \textnormal{\&} Triggers} to define and collect long-horizon signals (interaction + tool traces); \textit{Intervention Planning} to set targets (e.g., balancing capability expansion against thinking compression) and set rewards (e.g., evolving reward architectures from granular process feedback toward verifiable ground truth); and \textit{Training \textnormal{\&} Update Execution} to train under strict systemic consistency, eliminating algorithmic biases (e.g., difficulty or length bias). The resulting artifacts include a \textit{Reward-policy maturity evolution record}, \textit{Long-horizon signals and logged tool traces}, and \textit{Technical reports, with particular emphasis on continual RL details}. Overall, this guideline operationalizes trustworthy continual RL via monitored long-horizon signals, explicit reward/target planning, and consistency-constrained training to mitigate long-horizon drift.

\paragraph{\textbf{Guideline 4}: Iterative recipe optimization with online adaptation}

In the workflow illustrated in Fig.~\ref{fig:guidelines}(d), recipe optimization is carried out over update iterations across release cycles to improve cost/quality, addressing the gap of a manual recipe bottleneck. The mechanism concentrates on two components: \textit{Intervention Planning} to allocate update budget and choose automated recipe-optimization techniques under the target scenario and operational constraints; and \textit{Training \textnormal{\&} Update Execution} to optimize the initial training recipe and execute the update accordingly, while adaptively refining the training recipe on-the-fly during training. The resulting artifacts include \textit{versioned recipes for reuse}, a \textit{training-recipe change log and model capability change log}, and \textit{training reports with recipes}. Overall, this guideline operationalizes self-optimizing training recipes across update iterations to reduce manual recipe bottlenecks while tracking recipe and capability changes in a versioned manner.

\paragraph{\textbf{Guideline 5}: Accountability substrate for cascading updates in a versioned ecosystem}

As illustrated in Fig.~\ref{fig:guidelines}(e), P5 sits beneath P1–P4 and cuts across the entire pipeline, aggregating the artifacts produced by the other principles into a release-grade evidence and decision loop that supports governance over release/rollback. Upon completing P5, the workflow can return to P2 to begin the next update cycle. Concretely, the mechanism proceeds as follows: (i) an explicit release decision is made under risk criteria; (ii) an evidence package consolidates evaluations, budgets, signals, regressions, and diagnostics; (iii) delta-oriented documentation provides a change log and its expected impact; and (iv) external interfaces support selective transparency and incident learning to enable reliable release and rollback. Overall, P5 upgrades continual learning from ``continual training'' to \emph{governed continual release}: each update must leave reviewable evidence, pass interpretable gates, and preserve a viable rollback path—providing a system-level trust substrate for the mechanisms in P1–P4.

\subsection{Toward a research roadmap for ICL}

\begin{figure}[t]
  \centering
  \includegraphics[width=1.0\textwidth]{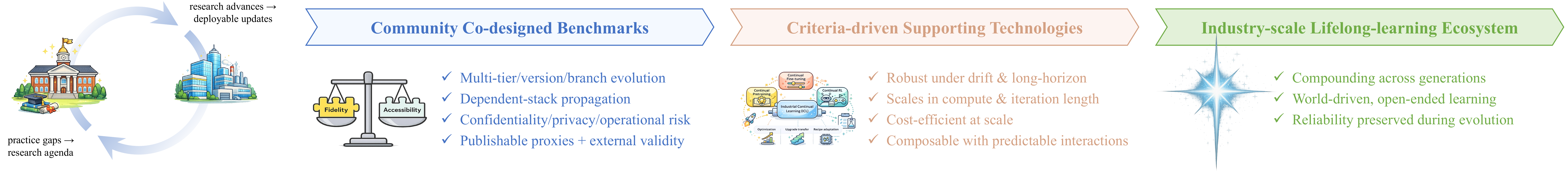}
  \caption{An illustrative roadmap: from co-designed benchmarks to criteria-driven techniques and an industry-scale lifelong-learning ecosystem.}
  \label{fig:icl-roadmap}
\end{figure}

In the previous section, we proposed a principle-driven, closed-loop set of guidelines for ICL. 
In this section, as shown in  Fig.~\ref{fig:icl-roadmap}, we further discuss how to feed the observable realities and critical gaps from industrial practice back to the research community, so as to catalyze academic progress and create a virtuous cycle through long-horizon iteration between industry and academia.

\paragraph{\textbf{Benchmark}: Community co-design to bridge fidelity and accessibility}

Current continual learning benchmarks~\cite{wang2023trace,zheng2024learn} in academia primarily target sequential learning over tasks for a single model (e.g., continual pretraining, continual fine-tuning, or continual preference alignment, as illustrated on the right side of Fig.~\ref{fig:lifecycle}). They are often not designed to capture the multi-tier, multi-version, and multi-branch evolution structures that dominate industrial deployments, where updates propagate across a dependent stack and are constrained by confidentiality, privacy, and operational risk. A key research direction is therefore community-level benchmark co-design with industry: to abstract and reconstruct the industrial technical problems with publishable proxy settings, so that methods can be compared openly and iterated on at substantially lower cost than full industrial-scale runs. Importantly, the central methodological challenge is external validity: benchmarks must balance (i) \emph{fidelity} to industrial dynamics with (ii) \emph{accessibility} under typical academic compute budgets. If the balance is not achieved, benchmarks may either yield conclusions that do not transfer when scaling to industrial regimes, or become too resource-intensive for the research community to iterate on---both of which would undermine their role as a catalyst for sustained progress.

\paragraph{\textbf{Technology}: Criteria-driven development of scalable supporting techniques}

Beyond any single bottleneck, ICL depends on a broader set of supporting technologies aligned with the taxonomy in Section~\ref{subsec:current_status} (spanning optimization, upgrade transfer, continual RL, and recipe adaptation). The research community can strengthen this space by prioritizing techniques that remain effective under the realistic benchmark settings above, rather than only under idealized assumptions. Concretely, a practical agenda is to evaluate and improve supporting techniques along four criteria: (i) \emph{robustness} under distribution drift, long-horizon update sequences, and heterogeneous objectives; (ii) \emph{scalability} in both compute and iteration length, such that gains persist as settings move toward industrial regimes; (iii) \emph{cost-efficiency}, so improvements do not rely on disproportionately expensive procedures that are unlikely to be adopted at scale; and (iv) \emph{composability}, i.e., predictable interaction effects when combined with other high-leverage mechanisms already used in practice, avoiding negative interference with more cost-effective components. The desired outcome is not a single ``best'' algorithm, but a set of reliable, modular building blocks whose scaling behavior, interaction patterns, and failure modes can be characterized on ecosystem-relevant benchmarks and then integrated into iterative update loops.

\paragraph{\textbf{North-star outlook}: Industry-scale lifelong-learning ecosystem}

Building on the ecosystem view of LLM evolution developed throughout this article, we use ``lifelong-learning ecosystem'' as an aspirational endpoint of ecosystem-level evolution: progress continues over long horizons without repeatedly resetting when base models, objectives, or deployment contexts change. A north-star vision is to make evolution itself a first-class, cumulative capability of the ecosystem: (i) improvements \emph{compound across generations} rather than being re-earned from scratch; (ii) learning becomes \emph{world-driven and open-ended} at the ecosystem level, reflecting evolving users, tools, and environments rather than a fixed task list; and (iii) reliability is preserved as the ecosystem evolves, so long-horizon progress does not trade off against systemic brittleness. Under this framing, ICL provides the enabling discipline that can turn ecosystem evolution from ad hoc iteration into a sustainable, comparable, and eventually compounding process, with the above benchmarking and composable supporting technologies serving as practical stepping stones toward that endpoint.

\section{Conclusion}
\label{conclusion}

The central claim of this paper is that ICL for LLMs should no longer be treated as single-model, anti-forgetting optimization. Instead, it should be viewed as a closed-loop lifecycle system whose updates propagate across layers, versions, and branches. Grounded in the real evolution dynamics of a versioned ecosystem, we distill three long-term challenges and systematize them into five design principles and a practical update-and-release loop.
Beyond summarizing current practice, this viewpoint offers a scalable way to organize and govern how AI systems evolve across many releases. As foundation models undergo more frequent version shifts, agentic applications depend increasingly on online interaction feedback, and compliance and safety requirements continue to rise, AI systems will resemble long-running critical infrastructure.  Competitiveness will therefore depend less on peak performance from a single training run and more on the ability to iterate under constraints, preserve and transfer capabilities across upgrades, and improve in a transparent, auditable, and reversible manner.

\bibliographystyle{ACM-Reference-Format}
\bibliography{main}

\end{document}